%% file: main.tex
\documentclass[10pt,twocolumn,letterpaper]{article}

\usepackage[pagenumbers]{cvpr} 

\usepackage{graphicx}
\usepackage{amsmath}
\usepackage{amssymb}
\usepackage{comment}
\usepackage{booktabs}
\usepackage{multirow}
\usepackage{enumitem}
\usepackage{makecell}
\usepackage{pifont}

\newcommand{\cmark}{\ding{51}}%
\newcommand{\xmark}{\ding{55}}%

\newcommand{\tinytit}[1]{\noindent\textbf{#1.}}
\newcommand{\quotationmarks}[1]{``#1''}

\usepackage[pagebackref,breaklinks,colorlinks]{hyperref}

\def\cvprPaperID{2286} 
\def\confName{CVPR}
\def\confYear{2022}

\begin{document}

\title{How many Observations are Enough?\\Knowledge Distillation for Trajectory Forecasting}

\author{
Alessio Monti$^1$ \qquad Angelo Porrello$^1$ \qquad Simone Calderara$^1$ \qquad Pasquale Coscia$^2$ \\ Lamberto Ballan$^2$ \qquad Rita Cucchiara$^1$\\\\
\begin{tabular}{cc}
\makecell{$^1$University of Modena and Reggio Emilia, Italy} & \makecell{$^2$University of Padova, Italy}\\
\end{tabular}
}

\maketitle

\input{sections/0_abstract.tex}

\input{sections/1_introduction}
\input{sections/2_related}
\input{sections/3_method}
\input{sections/4_experiments}
\input{sections/5_conclusion}

{\small
\bibliographystyle{ieee_fullname}
\bibliography{egbib}
}

\end{document}


\title{How many Observations are Enough?\\Knowledge Distillation for Trajectory Forecasting - Supplementary Material}

\author{
Alessio Monti$^1$ \qquad Angelo Porrello$^1$ \qquad Simone Calderara$^1$ \qquad Pasquale Coscia$^2$ \\ Lamberto Ballan$^2$ \qquad Rita Cucchiara$^1$\\\\
\begin{tabular}{cc}
\makecell{$^1$University of Modena and Reggio Emilia, Italy} & \makecell{$^2$University of Padova, Italy}\\
\end{tabular}
}

\maketitle

\section{Time lags}
\begin{table}[t]
\centering
\begin{tabular}{lccc}
\toprule
 & \textbf{$\nabla x_t = 1$} & \textbf{$\nabla x_t = 2$} & \textbf{$\nabla x_t = 3$} \\
Input & ($x_7, x_8$) & ($x_6, x_8$) & ($x_5, x_8$) \\
\midrule
\textbf{STT} & 0.73 / 1.44 & 0.80 / 1.53 & 0.88 / 1.63 \\
\textbf{DTO} & \textbf{0.64} / \textbf{1.27} & \textbf{0.66} / \textbf{1.32} & \textbf{0.69} / \textbf{1.37} \\
\bottomrule
\end{tabular}
\caption{On the SDD, an analysis of increasing time-lags $\nabla x_t$ with a fixed number of observed time steps (only two).}
\label{tab:timelagexp}
\end{table}
We also investigate how our method performs when increasing the lag between the two observation used for prediction. As highlighted in Tab.~\ref{tab:timelagexp}, the procedure we set up yields robust performance even in this setting.
\section{On the \quotationmarks{length-shift problem} -- additional considerations}
\begin{figure}
    \centering
    \includegraphics[width=\columnwidth]{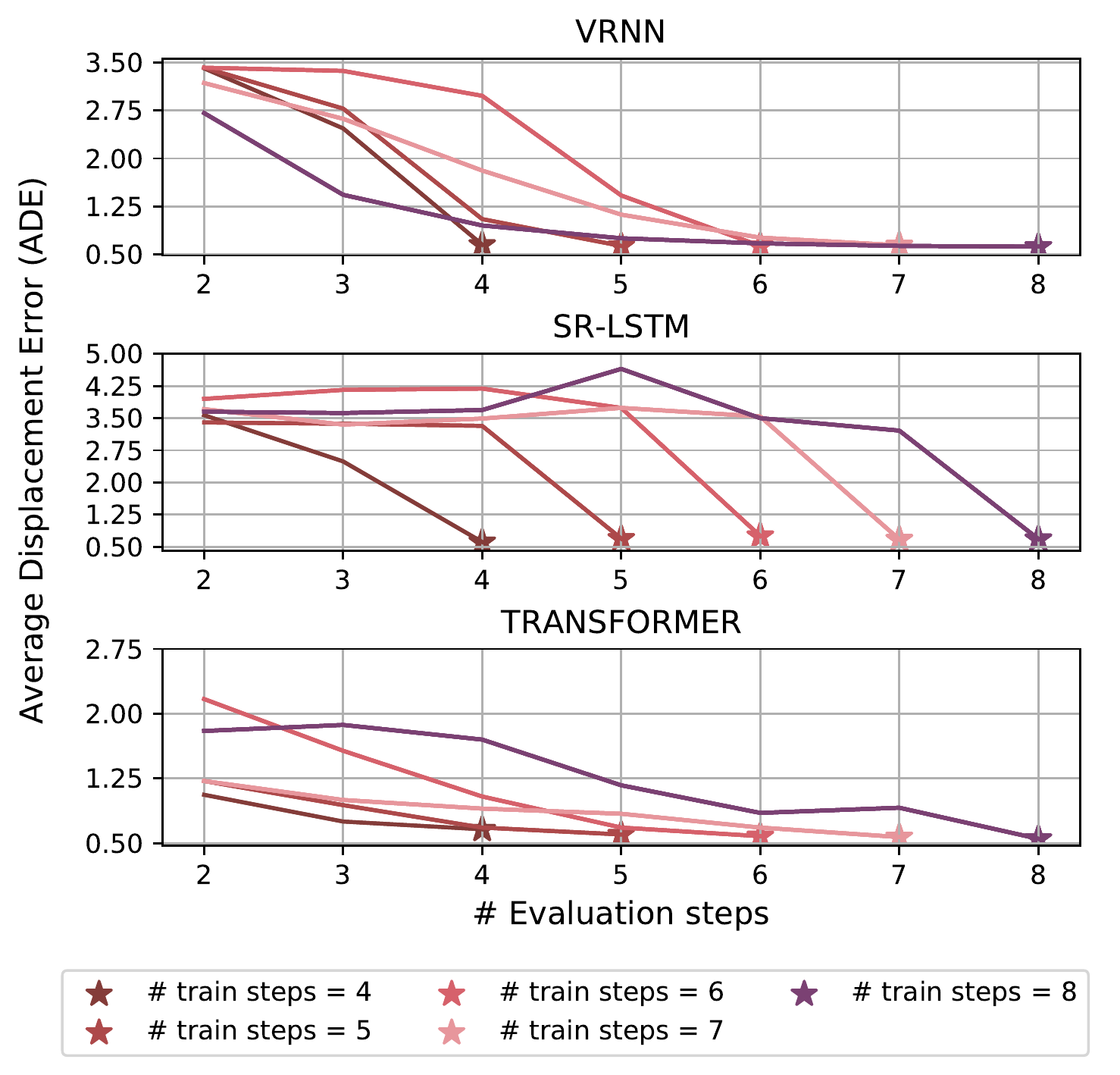}
    \caption{For different architectures (Variational RNN~\cite{chung2015recurrent}, SR-LSTM~\cite{zhang2019sr} and our STT), the performance trends when altering the number of observation time steps at evaluation time (ETH).} 
    \label{fig:lengtshift}
\end{figure}
As outlined in Sec.~5.5, trajectory prediction models show remarkable performance when evaluated according to the standard protocol, \emph{viz.}, $8$ observation time steps and $12$ prediction time steps. Nevertheless, we argue that their predictions overly bind to the data used at training time. To prove our intuition, we thoroughly investigate how models behave when the number of input time steps changes at evaluation time, which we define as \quotationmarks{\textbf{length-shift problem}}. 

As shown in Fig.~\ref{fig:lengtshift}, this behaviour is a common trait among different models: more specifically, we report the ADE metric for different categories of predictors (generative~\cite{chung2015recurrent}, LSTM-based~\cite{zhang2019sr}, attention-based) and observation-prediction splits ($8$--$12$, $7$--$12$, etc.). As expected, the optimum always occurs when train/test conditions met; on the contrary, altering the amount of input information (even slightly, \ie, removing a single time step) produces detrimental effects and delivers unacceptable inference errors. This behaviour could limit the usage of these models when there is no perfect match between train/test conditions.
\begin{table*}[t]
    \centering
    \small
    \begin{tabular}{llcccccccccccccc}
      \toprule
        \textbf{Dataset} & \textbf{Training strategy} & \textbf{obs=2} & \textbf{obs=3} & \textbf{obs=4} & \textbf{obs=5} & \textbf{obs=6} & \textbf{obs=7} & \textbf{obs=8} \\
        \midrule
        \addlinespace
        \multicolumn{1}{c}{\multirow{4}[4]{*}{SDD}} & From scratch & 0.73/1.44 & 0.67/1.31 & 0.65/1.30 & 0.65/1.29 & 0.64/1.29 & 0.65/1.32 & \underline{\textit{\textbf{0.63}/\textbf{1.26}}} \\
        \cmidrule{2-9}
        & Variable observations & 0.92/1.78 & 0.95/1.75 & 0.97/1.75 & 0.89/1.63 & 0.82/1.55 & 0.81/1.54 & 0.83/1.56 \\
        & Past generation & 0.85/1.58 & 0.76/1.45 & 0.73/1.42 & 0.74/1.45 & 0.71/1.39 & 0.67/1.35 & - \\
        & Distilling the Observations & \textbf{0.64}/\textbf{1.27} & \textbf{0.64}/\textbf{1.27} & \textbf{0.64}/\textbf{1.28} & \textbf{0.64}/\textbf{1.28} & \textbf{0.63}/\textbf{1.26} & \textbf{0.62}/\textbf{1.22} & 0.63/1.27 \\
        \addlinespace
        \bottomrule
    \end{tabular}
    \caption{Comparison (ADE/FDE) between different training strategies on the Stanford Drone Dataset; all methods are trained and tested on the same number of time steps, reported in the header. Best results are in bold. The distillation teacher is in underlined italic.}
    \label{tab:sdd_addr_length_shift}
\end{table*}

\medskip
\tinytit{Addressing the length-shift problem on the Stanford Drone Dataset} We also report our results using different training strategies on the Stanford Drone Dataset (Tab.~\ref{tab:sdd_addr_length_shift}). These results are similar to what has been observed on ETH/UCY and Lyft datasets. Training from scratch gives unsatisfactory results: the student is not able to observe enough information about agents' motion history. Training with a variable number of observations makes our model invariant to the amount of information it is fed with, but it delivers overly high errors. Generating missing observations introduces too much noise to obtain reliable predictions. As reported in the main paper, DTO appears the most promising strategy, being able to exploit enough past information even when the number of observations is limited.
\section{Spatial masking}
\begin{figure}
    \centering
    \includegraphics[width=\columnwidth]{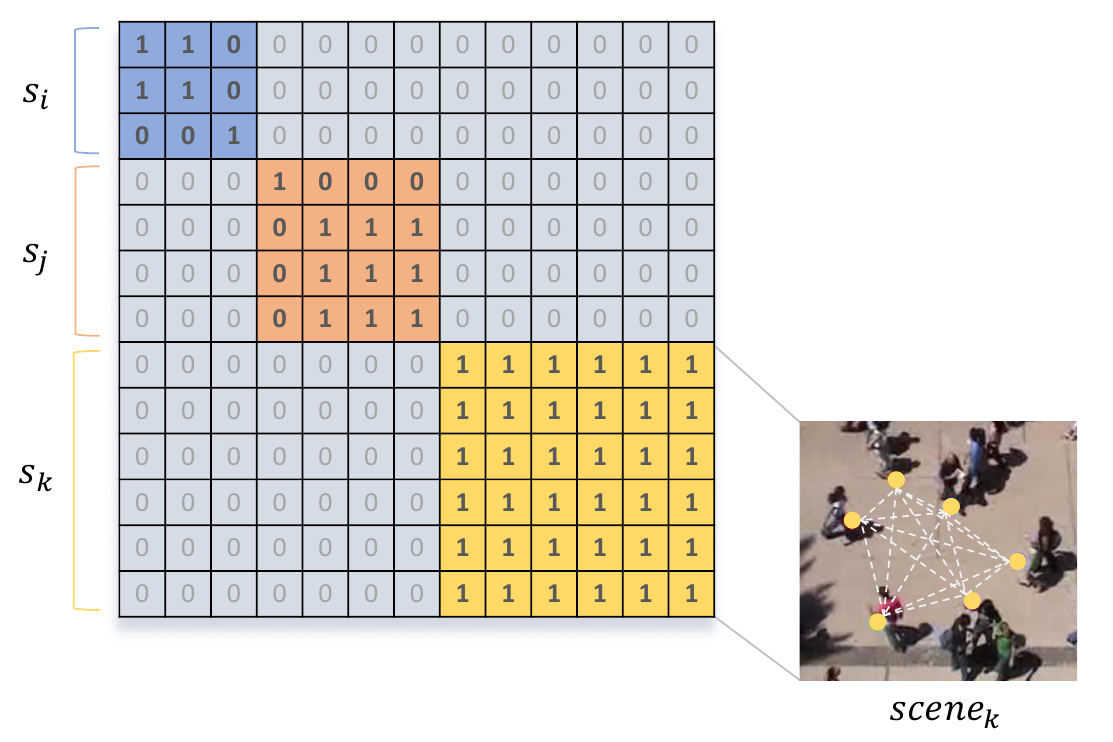}
    \caption{Example of a batch mask for the STT spatial attention. Before the softmax layer, logits corresponding to $0$ values are masked and set to $-\infty$. This way, agents are allowed to attend only on the neighbouring agents in the same scene (\eg scene $k$) at the given time step.} 
    \label{fig:spatial_masking}
\end{figure}
Our Spatio-Temporal solution (STT) relies on a single transformer-based architecture with shared parameters. Since batches are composed of randomly sampled scenes, performing a plain spatial self-attention without further solutions would result in attending on agents that appear in different scenes. To prevent attending on wrong agents, we build a dedicated spatial attention mask.

As done in~\cite{huang2019stgat,kosaraju2019social,zhang2019sr,yu2020spatio}, we represent the pedestrian space as an undirected graph. In doing so, the adjacency matrix describes spatial relationships occurring among different nodes (agents). We employ a distance-based adjacency matrix where, in order to focus only on the neighbourhood, we preserve (\ie, set to 1) the edges that are associated to distances smaller than a given threshold $d$. Moreover, as shown in Fig.~\ref{fig:spatial_masking}, we cluster adjacency matrices of different scenes in batches by devising a block irregular matrix: this matrix is built by aligning several adjacency matrices on the main diagonal and by setting to $0$ the remaining values.

By employing this matrix as a mask applied to the logits before the softmax operation, we are able to restrict the attention scope only to agents that: \textit{i)} share the same scene and time steps; \textit{ii)} are spatially close to each other. In our implementation, we set to $-\infty$ all the attention logits corresponding to $0$-values in the mask.
\section{Datasets preprocessing}
\tit{Frame-rate} Following the ETH/UCY setting proposed in~\cite{alahi2016social,gupta2018social, pellegrini2009you}, we employ a sampling rate of 2.5 FPS. By sampling agents' positions every $0.4s$, employing $8$ frames of observation and $12$ frames of prediction corresponds to observe the trajectories for $3.2~s$ and predicting their future development in the next $4.2~s$. We use the same frame also for both Stanford Drone and Lyft datasets.
\tit{Data preparation} Several previous works~\cite{alahi2016social,gupta2018social,kosaraju2019social,amirian2019social} represent input trajectories as a series of relative displacements: namely, each absolute position $\mathbf{x}_t$ is transformed in a couple of displacements $(\Delta x_t, \Delta y_t)$ w.r.t. the previous position $\mathbf{x}_{t-1}$. Differently,~\cite{zhang2019sr} proposes to preserve absolute positions and normalize them by subtracting the last observation: this procedure, referred to as \emph{Nabs}, seems to grant higher benefits. We opt for this solution (as in~\cite{yu2020spatio}).
\tit{ETH/UCY issues} Recent works~\cite{zhang2019sr, scholler2020cvm} brought up several issues affecting the ETH/UCY dataset.~\cite{zhang2019sr} points out that the original video used to obtain the labelled trajectories for the ETH scenario is accelerated, thus strongly affecting the resulting motion patterns: when sampling at a fixed rate, the trajectories of this scenario present higher speeds than the ones captured in the remaining scenes. The authors mitigate this issue by treating $0.4$ s as $6$ frames instead of the original $10$ frames. To ensure a fair comparison~\cite{zhang2019sr, yu2020spatio}, we choose to adopt the same correction. 

In addition,~\cite{scholler2020cvm} shows that Hotel scene mostly contains trajectories that are orthogonal to the ones contained in the other four scenarios. This peculiarity could prevent the model from learning useful environmental priors. To overcome this issue, we follow~\cite{zhang2019sr,yu2020spatio} and employ data augmentation by applying a random rotation to each position inside each mini-batch.

\section{Implementation details}
\tinytit{Teacher} We initialize the weights of our architecture according to~\cite{glorot2010understanding}. All teacher networks are trained for $1000$ epochs using Adam~\cite{kingma2015adam} as optimizer.

For ETH/UCY, our Spatio-Temporal Transformer employs two layers for each encoder and decoder stack. We use internal embeddings of size $d_{\text{model}}=64$ and set the dimension $d_{\text{ff}}$ of the position-wise feed-forward inner layer to $128$. We employ $8$ attention heads for each temporal and spatial attention and set the size of queries, keys and values to $d_{\text{k}} = d_{\text{v}} = d_{\text{model}}/8 = 8$:
reducing the dimension of each head allows to obtain a similar computation cost to a single-head attention with \quotationmarks{full} dimensionality \cite{vaswani2017attention}. We use a batch size of $16$ and a learning rate of $10^{-4}$.

For SDD, our Spatio-Temporal Transformer employs a single layer for each stack. We use internal embeddings of size $d_{\text{model}}=32$ and set the dimension $d_{\text{ff}}$ of the position-wise feed-forward inner layer to $128$. We employ $8$ attention heads for each temporal and spatial attention and, as reported above, obtain the size of queries, keys and values by dividing $d_{\text{model}}$ by the number of heads. We use a batch size of $32$ and a learning rate of $5 \cdot 10^{-4}$.

For Lyft, our Spatio-Temporal Transformer employs a single layer for each stack. We use internal embeddings of size $d_{\text{model}}=32$ and set the dimension $d_{\text{ff}}$ of the position-wise feed-forward inner layer to $128$. We employ $8$ attention heads for each temporal and spatial attention and, as said above, obtain the size of queries, keys and values by dividing $d_{\text{model}}$ by the number of heads. We use a batch size of $32$ and a learning rate of $5 \cdot 10^{-5}$.

\tinytit{Distillation} Regarding the student initialization, we empirically found more beneficial to inherit the teacher weights rather starting from scratch. We believe that starting from a solid parametrization eases the student effort: this way, the student can just adjust the previously learned weights while preserving as much knowledge as possible. In contrast, we observe that learning weights from scratch represents an overly detrimental situation: the student rarely approaches the results of its teacher. 

Furthermore, we set our teacher network in training mode during distillation. In this way, the statistics of the different normalization layers are computed on a batch basis: as reported in~\cite{bagherinezhad2018label}, this grants more accurate teacher supervision.
{\small
\bibliographystyle{ieee_fullname}
\bibliography{egbib}
}

%% file: sections/0_abstract.tex
\begin{abstract}
Accurate prediction of future human positions is an essential task for modern video-surveillance systems. Current state-of-the-art models usually rely on a \quotationmarks{history} of past tracked locations (\textit{e.g.}, 3 to 5 seconds) to predict a plausible sequence of future locations (\textit{e.g.} up to the next 5 seconds). We feel that this common schema neglects critical traits of realistic applications: as the collection of input trajectories involves machine perception (\textit{i.e.}, detection and tracking), incorrect detection and fragmentation errors may accumulate in crowded scenes, leading to tracking drifts. On this account, the model would be fed with corrupted and noisy input data, thus fatally affecting its prediction performance.

In this regard, we focus on delivering accurate predictions when only few input observations are used, thus potentially lowering the risks associated with automatic perception. To this end, we conceive a novel distillation strategy that allows a knowledge transfer from a teacher network to a student one, the latter fed with fewer observations (just two ones). We show that a properly defined teacher supervision allows a student network to perform comparably to state-of-the-art approaches that demand more observations. Besides, extensive experiments on common trajectory forecasting datasets highlight that our student network better generalizes to unseen scenarios.
\end{abstract}

%% file: sections/1_introduction.tex
\section{Introduction}
\label{sec:introduction}
Pedestrian trajectory forecasting deals with predicting future paths through the exploitation of individual trajectory information and mutual influence between pedestrians. This task has several practical applications in advanced surveillance systems~\cite{9373939}, behavioral analysis~\cite{rudenko2020human}, intrusion detection~\cite{13413837}, smart vehicles and autonomous systems~\cite{9043898,bartoli2018context}.

While several recent works focused on novel deep-network architectures tailored for this task~\cite{gupta2018social, huang2019stgat, zhang2019sr, giuliari2020transformer, trajectronpp, yuan2021agentformer}, we believe the inference phase of a trajectory predictor has not been thoroughly addressed and investigated yet. Typically, data-driven models are trained and evaluated on large public datasets of tracked trajectories refined with a human intervention to correct missed detections and identity switches. However, this process is unfeasible at inference time: therefore, the input trajectories required to condition the prediction have to be automatically extracted by a tracking system.

In this regard, the widely adopted $8$-$12$ protocol~\cite{alahi2016social, gupta2018social, Sadeghian_2019_CVPR, huang2019stgat, zhang2019sr, kosaraju2019social, yu2020spatio}~(\textit{i.e.}, $8$ input time steps and $12$ ones for prediction), which require data collected at $2.5$ FPS, does not provide a large margin of correction for the above situations. In real-time applications, a visual tracking system may provide inaccurate observations for sequences of such length~\cite{MOTChallenge20}: occlusions, false detections and non-rigid shape deformations pose non-trivial issues.

To overcome the aforementioned limitations, one potential solution is to reduce the length of the input trajectories to the extent we can minimize the tracking associations errors. Based on this intuition, this work proposes an approach based on Knowledge Distillation~\cite{hinton2015distilling}, which recovers a reliable proxy of the same information obtained with more input observations. We show that it allows for an effective inference schema, requiring fewer samples than the training one. We also demonstrate that properly conditioning a model on shorter input trajectories provides more room to generalize across different experimental settings.

From a technical perspective, this work implements our idea by deploying a teacher-student paradigm~\cite{hinton2015distilling}: a student network is trained to mimic a teacher's behaviour using less input observations. Each network devises a transformer-based architecture that accounts for both \emph{spatial} and \emph{temporal} interactions through an attention mechanism. To deal with a limited number of observations, we propose a distillation procedure that acts on both encoder and decoder stacks of the transformer architecture. Finally, our objective function takes into account ground-truth data and distillation losses to conveniently match teacher and student internal representations.

We remark the following contributions: \textit{i)} to the best of our knowledge, this is the first attempt to in-depth analyze the effectiveness (at inference time) of the evaluation protocol commonly employed by current trajectory prediction models; \textit{ii)} we introduce a novel distillation strategy to reduce the length of the input trajectories while keeping the prediction accurate; \textit{iii)} we explore the student's capabilities in adapting and transferring its knowledge to scenarios exhibiting different levels of complexity in terms of human dynamics and scene interactions.
Indeed the experiments highlights that is possible to construct a solid trajectory forecasting system that at inference time is fed only with just 2 observation (i.e. last two) per-pedestrian. This is possible only through the thoughtful exploitation of the global knowledge that can be inferred from training data and distilled into the inference model.

%% file: sections/2_related.tex
\section{Related Work}
\tinytit{Social models} Modelling human-human interactions plays a fundamental part in predicting plausible trajectories. Pioneering works take advantage of hand-crafted relations, energy-based features or rule-based models~\cite{helbing1995social, treuille2006continuum, antonini2006discrete, wang2006gaussian, pellegrini2009you}, which fail to adapt to scene changes and model complex crowd dynamics.
In recent years, data-driven approaches received increasing attention: in their seminal work, Alahi \etal~\cite{alahi2016social} capture these interactions by aggregating the hidden states of neighbouring agents with a dedicated grid-based ``Social Pooling''.
Gupta \etal~\cite{gupta2018social} improved this mechanisms which is extended to all the agents involved in the scene by max-pooling their hidden states. Such modules have also been extended with attention-based mechanisms~\cite{amirian2019social}, while other works propose architectures combining social pooling with context information (\eg scene semantic, groups or head poses)~\cite{hasan2018mx,bartoli2018context,bisagno2018group,lisotto2019social,Sadeghian_2019_CVPR}.

Recent improvements in graph machine learning~\cite{kipf2016semi,velivckovic2017graph,li2016gated,porrello2019classifying} motivated the adoption of such flexible structures to model agents relationships.
Several solutions~\cite{vemula2018social,kosaraju2019social,huang2019stgat,sun2019stochastic,sun2020recursive, bertugli2021} consider agents as graph nodes whose features are represented by their hidden states. This solution enables the use of message-passing mechanisms and grants the possibility to aggregate information at each node with powerful Graph Neural Networks (GNNs) like Graph Attention Networks~\cite{velivckovic2017graph}.
Zhang \etal~\cite{zhang2019sr} similarly treat the pedestrian space as a fully-connected graph but design a custom message-passing solution that integrates a \textit{motion gate} to perform a feature selection based on pedestrian movements.
Finally, Yu \etal~\cite{yu2020spatio} only rely on attention mechanisms to predict future locations exploiting recent advances in transformer-based architectures~\cite{vaswani2017attention}.

\medskip
\tinytit{Knowledge distillation} Knowledge distillation has been firstly investigated as an approach for model compression~\cite{bucilu2006model,hinton2015distilling}: a small model (\emph{student}) has to mimic the behaviour of an over-parameterized one (\emph{teacher}). As a result, the student manifest a smaller memory footprint without experiencing a large drop in the overall performance. \cite{romero2014fitnets} aims to reduce both student and teacher feature maps; ~\cite{hinton2015distilling} suggests to match the soft-targets before the final classification layer;~\cite{zagoruyko2017payingmoreattention} matches the features of attention regions.

In this work, we employ knowledge distillation in a different fashion. Inspired by~\cite{furlanello2018born,zhang2019beyourown}, our aim is not to compress a model yet to improve its performance. This procedure is usually referred to as \textit{self-distillation}, since the student network shares the same architecture of its teacher. Similarly to~\cite{bhardwaj2019efficient,porrello2020robust}, our approach sets up asymmetric networks: the student is encouraged to overcome its knowledge gap by following the guide of its teacher, eventually boosting its performance. This is done in the specific context of trajectory forecasting, and we demonstrate that knowledge distillation can lead to effective predictions even when the model has access to very few observations.

%% file: sections/3_method.tex
\section{Model}
Trajectory forecasting is usually defined as a time-series prediction problem~\cite{rudenko2020human}. The task is particularly challenging because: \emph{i)} human motion is inherently multi-modal, and \emph{ii)} agents simultaneously interact with each other and with static scene elements.

To meet these two points, we design 
a novel approach modelling both temporal and spatial relationships occurring between agents. Specifically, this section describes how we extend the original Transformer~\cite{vaswani2017attention} architecture to deal with trajectory forecasting.

\subsection{Vanilla Transformer for Trajectory Prediction}
To deal with sequence-to-sequence tasks, transformers follow the well established encoder-decoder paradigm. Instead of relying on internal recurrent layers, input sequences are processed as a whole through a purely attentive mechanism. Self-attention aims to discover relationships between every pair of elements in the sequence: this reduces the risk of forgetting past information and allows the network to learn long-range dependencies~\cite{vaswani2017attention}.

From a technical perspective, each embedding $e_t$ (from time step $t=1$ up to $t=T$) is linearly projected into a triplet of vectors: a query \textit{q}$_t$, a key \textit{k}$_t$ and a value \textit{v}$_t$. Then, transformers exploit the dot product between queries and keys to compute attention coefficients (\textit{scaled dot-product attention}), the latter being used to weight the corresponding values and provide the final output. This operation is performed $h$ times (\textit{heads}) on different linear projections of $Q$, $K$ and $V$ to attend information from several representations at different positions. %
\subsection{Spatio-Temporal Transformer (STT)}
\label{sec:stt}
\begin{figure}
    \centering
    \includegraphics[width=0.99\columnwidth]{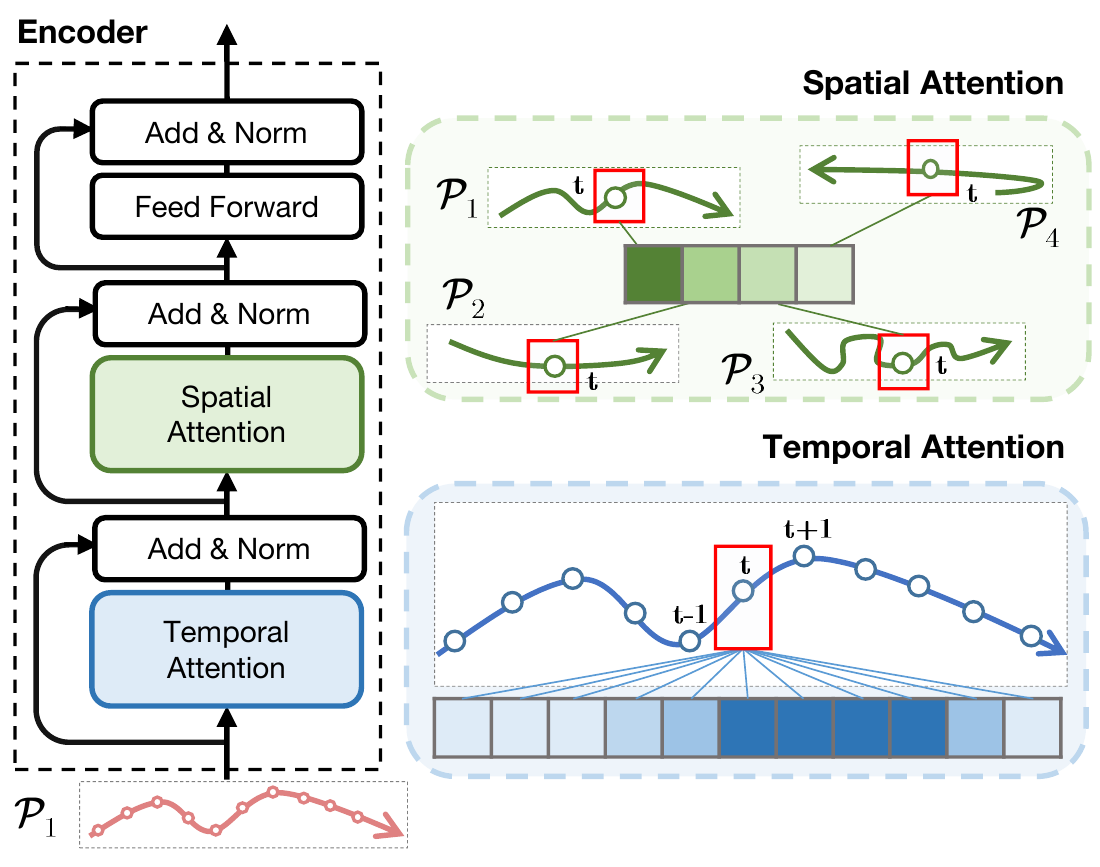}
    \caption{Our \textit{spatio-temporal} attention module. A \textit{temporal} encoder exploits temporal relationships between subsequent time steps of input sequences while a \textit{spatial} encoder collects human-human interactions that occur between agents at a fixed time step.}
    \label{fig:encoder_transformer}
\end{figure}

As in~\cite{vaswani2017attention}, our proposal firstly attends on the temporal axis with a scaled dot-product attention module. This way, it is able to recover temporal dependencies across different time steps and capture characteristic motion patterns of the monitored agents. However, this vanilla sequence-to-sequence model does not explicitly account for high-level spatio-temporal structure, i.e., no interactions are considered. For this reason, the output of our temporal attention is fed to a second self-attention module that acts on the spatial axis. While in the temporal attention queries, keys and values refer to different time steps of a specific agent, here \textit{Q}, \textit{K} and \textit{V} refer to the embeddings of all the agents at a fixed time step. This way, each agent can additionally attend on the information of its neighbours, recovering useful spatial information. Fig.~\ref{fig:encoder_transformer} shows a visual representation of our encoder architecture (the same applies to the decoder).
\begin{figure*}[t]
    \centering
    \includegraphics[width=0.91\textwidth]{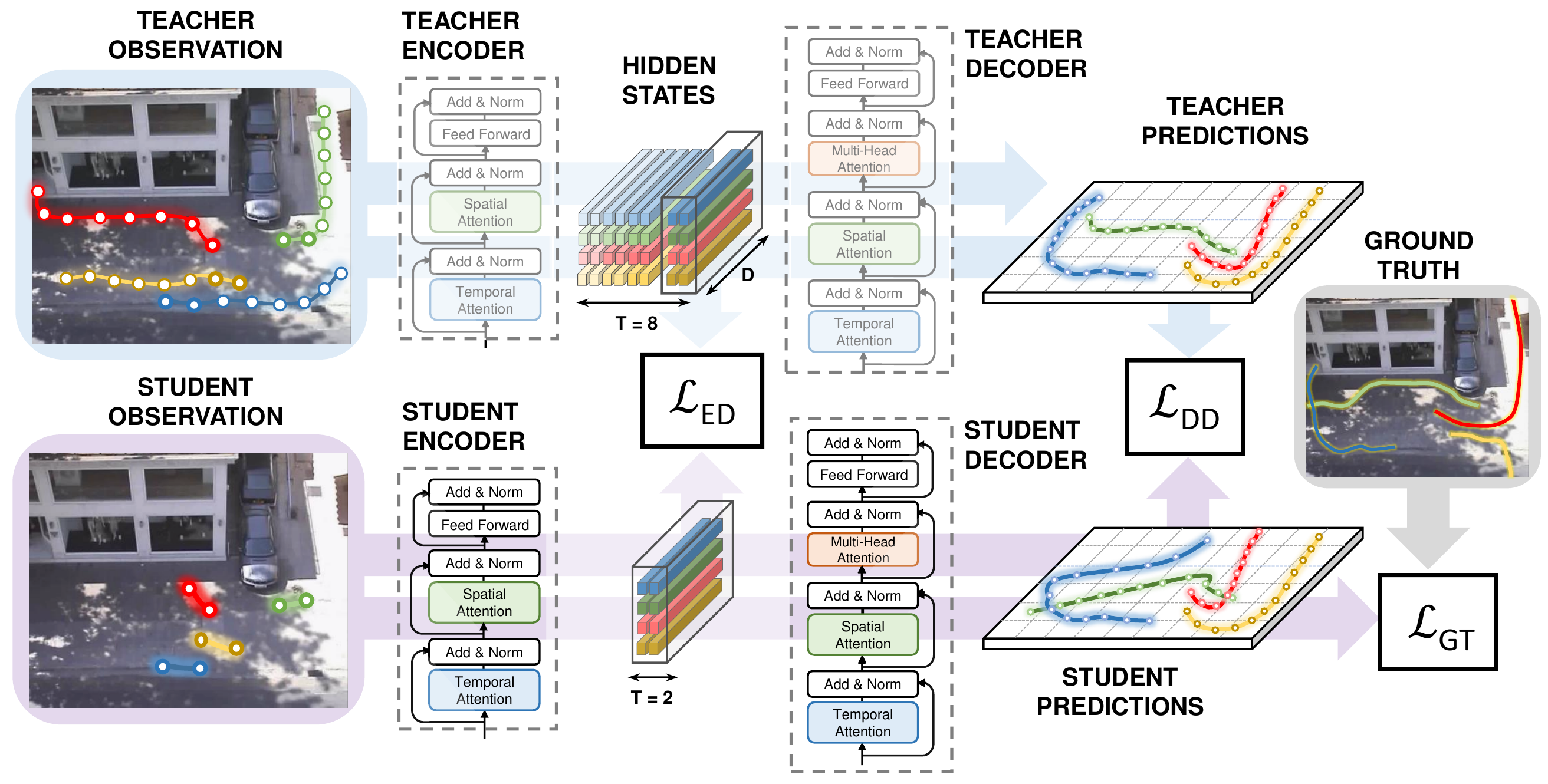}
    \caption{A comprehensive picture of our framework, termed Distilling the Observations (DTO), which provides a training strategy for obtaining accurate trajectory predictions when only few observations are available.}
    \vspace{-0.5em}
    \label{fig:framework}
\end{figure*}

\medskip
\tinytit{Relation with previous works} While the approach discussed in~\cite{giuliari2020transformer} consists of a transformer network treating each pedestrian separately (thus handling only temporal information), our self-attention mechanism takes into consideration also spatial interactions between pedestrians. This is somewhat similar to what has been devised by the authors of~\cite{yu2020spatio}, who equipped a spatio-temporal transformer with an auxiliary memory retaining representations of previous predictions. However, our approach significantly differs in the design of the decoder: while~\cite{yu2020spatio} adopts a fully connected layer, we stay close to the original transformer~\cite{vaswani2017attention} and mirror the encoder into the decoder.
\section{Distilling the Observations (DTO)}
Our goal is to set up a model capable of accurately predicting future positions when only a few observations are available: this way, we can address the inference-time shortcomings outlined in Sec.~\ref{sec:introduction}. More specifically, we devise a two-fold approach (depicted in Fig.~\ref{fig:framework}):
\begin{itemize}[noitemsep]
    \item \textit{firstly} (Sec.~\ref{sec:firststep}), we train a teacher network to estimate trajectories given $8$-length observation sequences;
    \item \textit{secondly} (Sec.~\ref{sec:secondstep}), we freeze its parameters and attempt to transfer its predictive capability to a student network. Importantly, the latter is forced to operate with an information gap, \ie, using only a small fraction of available inputs (\eg, two last observations).
\end{itemize}
\subsection{Teacher training}
\label{sec:firststep}
To train our teacher network, we follow the standard protocol and consider $8$ observation time steps and $12$ prediction time steps. The network is trained by \textit{teacher forcing}, \ie, when predicting the next time step, the decoder is conditioned on past ground-truth samples rather than its own predictions. We mark the beginning of the prediction sequence with a \textit{start} token and mask the information related to the future time steps. Mean Squared Error (MSE) between predictions and ground-truth positions is used as loss function while training our teacher network:
\begin{equation}
\label{eq:teacher_loss}
\mathcal{L}_{\text{GT}} = \frac{1}{P} \sum_{p=0}^{P-1} \left\lVert \mathbf{x}_{p, [:]} - \mathbf{\hat{x}}_{p, [:]} \right\rVert^2,
\end{equation}
where $P$ is the number of pedestrians and $\mathbf{x}_{p,[:]}$ ($\mathbf{\hat{x}}_{p,[:]}$) represents the sequence of ground-truth (predicted) positions of a pedestrian $p$ at time $t$.

By contrast, the inference procedure resembles an auto-regressive model. The decoder forecasts the first future position using the last hidden state of the encoder stack and an input sequence initially composed only by the \textit{start} token. At each step, the predicted position $\mathbf{\hat{x}}_{t}$ is concatenated to the current input sequence: this partial sequence is fed again to the decoder to predict the next position $\mathbf{\hat{x}}_{t+1}$.
\subsection{Student training}
\label{sec:secondstep}
To preserve teacher's predictive capabilities given only few observations, our training strategy relies on transferring the knowledge lying in the entire input sequence: to achieve this, we act on both encoder and decoder stacks.

\medskip
\tinytit{Encoder distillation} Firstly, we force the student encoder to mimic the behaviour of its teacher's counterpart. Given the information gap between the two networks, \textit{the higher} the transfer occurring at this level, \textit{the higher} the capability of the encoder to infer the missing information from the (few) spatio-temporal interactions it observes. Technically, we focus on the final hidden representations produced by the encoder stack (\ie, the outputs of the fully connected layer) and match those corresponding to the common time steps using the following loss: 
\begin{equation}
    \label{eq:encoder_distillation_loss}
    \mathcal{L}_{\text{ED}} = \frac{1}{P} \sum_{p=0}^{P-1} \left\lVert h^T_{p, [T-K:T]} -  h^S_{p, [0:K]} \right\rVert^2,
\end{equation}
where $h^T_{:, :}$ are the activations of the teacher encoder, $h^S_{:, :}$ are the student' ones, while $T$ and $K$ are the number of observations we feed to the teacher and the student, respectively.

\medskip
\tinytit{Decoder distillation} At the same time, we focus on matching the function space spanned by both teacher and student decoders. We pursue our goal relying on two terms: on one hand, we match the activations prior to the fully connected layer that give the final prediction, \ie $\mathbf{\hat{x}}_{p, t} = \text{FC}(o^T_{p, })$. On the other hand, as proposed in~\cite{wang2020minilm}, we exploit the self-attention coefficients $\textbf{A}^T_{p, [:]}$ of the last decoder layer as an additional learning guidance. The corresponding objective function is defined as follows:
\begin{equation}
    \label{eq:decoder_distillation_loss}
    \mathcal{L}_{\text{DD}} = \frac{1}{P} \sum_{p=0}^{P-1} \left\lVert o^T_{p, [:]} -  o^S_{p, [:]} \right\rVert^2 + \left\lVert \textbf{A}^T_{p, [:]} -  \textbf{A}^S_{p, [:]} \right\rVert^2.
\end{equation}
%
\tinytit{Overall objective} Finally, the student objective consists of a weighted sum of the prediction loss, which takes into account ground-truth positions, and the distillation losses:
\begin{equation}
    \label{eq:student_loss}
    \mathcal{L} = \alpha \mathcal{L}_{\text{GT}} + \beta \mathcal{L}_{\text{ED}} + \gamma \mathcal{L}_{\text{DD}},
\end{equation}
where $\alpha$, $\beta$ and $\gamma$ are three hyperparameters balancing the contribution of each term.

%% file: sections/4_experiments.tex
\section{Experiments}
\tinytit{Metrics} We consider two standard error metrics in our comparisons: the \textit{Average Displacement Error} (ADE) and the \textit{Final Displacement Error} (FDE)~\cite{pellegrini2009you}. While the ADE indicates the average Euclidean distance between all the predicted time steps and the ground-truth ones, the FDE expresses instead only the error regarding the final position.
\subsection{Datasets}
 
\tinytit{ETH/UCY} As usually done~\cite{alahi2016social, gupta2018social}, we stitch together two scenes from ETH~\cite{pellegrini2009you} (ETH and Hotel) and three scenes from UCY~\cite{Lerner2007CrowdsBE} (Univ, Zara-1, Zara-2). The resulting dataset contains more than $1500$ pedestrians, taking linear and non-linear paths in outdoor scenarios. We follow the common leave-one-scene-out protocol, training on $4$ scenes and testing on the remaining one.

\tinytit{Stanford Drone Dataset (SDD)}~\cite{sdd} is a large scale dataset collected by a drone monitoring crowded university campus scenarios. It contains multiple interacting agents (\eg, pedestrians, cyclists, cars) and is composed of a large diversity of urban scenes (\eg, intersections and parks) where people exhibit complex dynamics. We split the SDD World Plane Human-Human dataset~\cite{sadeghiankosaraju2018trajnet} into \textit{train} ($70~\%$), \textit{val} ($10~\%$) and \textit{test} ($20~\%$) sets, respectively.

\tinytit{Lyft Prediction Dataset}~\cite{lyft2020} is one of the largest collection of traffic agent motion data. It includes tracks of cars, pedestrians and other traffic agents recorded by cameras and lidar sensors of Lyft autonomous fleet. We split the reduced version of this dataset (1000 agents) in \textit{train} ($70~\%$), \textit{val} ($10~\%$) and \textit{test} ($20~\%$) sets.
\input{tables/results}
\subsection{Comparison with the State-of-the-art}
\label{sec:comparisons}
Since our proposal is deterministic (\ie, it gives a single future sample), we leave aside stochastic methods~\cite{gupta2018social, kosaraju2019social, Sadeghian_2019_CVPR} and compare our model to the following state-of-the-art deterministic solutions:
\begin{itemize}[noitemsep]
    \item Constant Velocity Model (CVM)~\cite{scholler2020cvm}: a simple but effective baseline that estimates future positions by considering solely the latest two timesteps;
    \item ST-GAT~\cite{huang2019stgat}: graph-based attention network using LSTMs to model temporal correlations. We consider the 1V-1 version, \ie without variety loss and with one output sample per input;
    \item Ind-TF~\cite{giuliari2020transformer}: vanilla transformer without explicit interactions modelling;
    \item SR-LSTM~\cite{zhang2019sr}: LSTM-based network integrating a system of motion gates that refines cells' hidden states using neighbourhood information;
    \item STAR~\cite{yu2020spatio}: encoder-decoder architecture based on a transformer network to model temporal information and spatial interactions. We consider its deterministic version obtained removing the Gaussian noise.
\end{itemize}
Tab.~\ref{tab:eth_adefde_sota_comparison} and Tab.~\ref{tab:sddlyft_adefde_sota_comparison} report our results: when trained according to the common protocol ($8$ observations -- $12$ predictions), our teacher network (STT -- 8 obs) performs comparably to state-of-the-art approaches. Notably, the student network (STT + DTO -- 2 obs, last row of Tab.~\ref{tab:eth_adefde_sota_comparison} and Tab.~\ref{tab:sddlyft_adefde_sota_comparison}) shows remarkable results: it approaches the teacher on all the datasets, suggesting that the last two observations are an informative summary of the input trajectory. Notably, trivial strategies using only two observations do not achieve the accuracy of our approach: both the CVM and training from scratch with short sequences (STT -- 2 obs) deliver higher errors. Instead, our training procedure successfully bridges the huge informative gap simulated at inference time.
\begin{table}
\centering
\begin{tabular}{lcc}
\toprule
	& \textbf{SDD} & \textbf{Lyft} \\
\midrule
	Ind-TF~\cite{giuliari2020transformer} & 0.74 / 1.46 & 0.31 / 0.62 \\
	CVM~\cite{scholler2020cvm} & 0.69 / 1.39 & 0.29 / 0.61 \\
	SR-LSTM~\cite{zhang2019sr} & 0.72 / 1.47 & \textbf{0.20} / \textbf{0.43} \\
\midrule
	\textbf{STT} (8 obs)& \textbf{0.63} / \textbf{1.26} & 0.24 / 0.53 \\
	\textbf{STT} (2 obs)& 0.73 / 1.44 & 0.31 / 0.56 \\
	\textbf{STT + DTO} (2 obs)& 0.64 / 1.27 & 0.27 / 0.55 \\
\bottomrule
\end{tabular}
\caption{ADE/FDE results on SDD and Lyft.}
\label{tab:sddlyft_adefde_sota_comparison}
\end{table}

It is worth noting that \textbf{two observations as input do not necessary generate straight lines as output}: in this case, our approach would have achieved results in line with those of the Constant Velocity Model (CVM) (which predicts straight lines by design). Instead, our results show that this does not happen: even observing two observations solely, indeed, our method takes also spatial relationships among pedestrians into account. This aspect is further corroborated by the superiority of our proposal w.r.t. Ind-TF, which treats every trajectory independently. In this regard, we argue that handling spatial interactions interacts well our distillation technique, closing the gap between using 8-samples input trajectories (Ind-TF) and 2-samples alone (STT -- 2 obs): the teacher may drive the student towards novel robust representations (\textit{e.g.} a better understanding of the spatial interactions within the local neighborhood).
\input{tables/vrnn}
\input{tables/tracking}
Finally, to support our choice of considering only deterministic methods, we rank multiple trajectories based on other criteria which do not require ground-truth annotations at inference time (\textit{e.g.}, for a V-RNN model, the aggregated KL-divergence between the approximate posterior and the conditional prior). In this regard, Tab.~\ref{tab:stochastic_vs_deterministic_tracked} proposes a comparison between our proposal and a V-RNN model with different sampling strategies: remarkably, only the (unviable) criterion based on ground-truth trajectories yields higher accuracy w.r.t.~DTO.
\subsection{Towards an ``in-the-wild'' evaluation: a tracker in the middle}
As outlined in Sec.~\ref{sec:introduction}, on-line scenarios cannot rely on human intervention to correct detection and re-identification errors at inference time. Based on this motivation, we advocate for employing shorter temporal horizons while estimating future trajectories ($2$ time steps in place of the common $8$ ones), since a tracker can still provide reliable predictions for so short fragments. To shed light on this point, we conduct an experiment on the Stanford Drone Dataset: more precisely, we focus on input trajectories and replace ground-truth associations with Deep SORT's~\cite{wojke2017simple} output, which is a tracking-by-detection algorithm that leverages a deep metric for modelling appearance. For each scene, we extract all the detections contained in the observation history; then, we run this tracker on the obtained detections and select as our new observation sequence the most similar tracklet to the ground-truth one. For the sake of simplicity, we restrict our analysis to examples that are successfully followed for at least 8 time steps (hence, we discard cases suffering from identity switches).

In this setting, we evaluate the performance of teacher (namely, STT fed with $8$-length tracklets) and student networks (namely, STT trained via DTO fed with $2$-length tracklets). As reported in Tab.~\ref{tab:gt_tracked_stt_vs_dto}, while DTO appears not advantageous in ideal scenarios (\textit{i.e.} using ground-truth observations), switching to a fully-automatic inference (\textit{i.e.} using tracked trajectories) turns the table: in almost all SDD scenes, our model trained via DTO experiences a diminished degradation in performance w.r.t. the teacher. Its worsening is mainly due to errors accumulation occurring on long sequences: as depicted in Fig.~\ref{fig:tracked_ades}, tracker's errors (measured as ADE between ground-truth and re-tracked input trajectories) spread on higher values for longer tracklets. By contrast, when restraining the model to just a few observations, the average association error tends to be lower, thus affecting less the downstream forecasting model.

Additionally, Tab.~\ref{tab:gt_tracked_stt_vs_dto} draws a comparison between DTO and two baselines that use only $2$ observations: STT trained from scratch with $2$ observations and the Constant Velocity Model (CVM). As reported, the effectiveness of DTO in a fully-automatic context is not merely due to the use of few time steps, but, more interestingly, also derives from the exploitation of our knowledge distillation paradigm.
\subsection{Why Distilling the Observations works}
\begin{figure}
  \centering
  \subfloat[]{\includegraphics[scale=0.323]{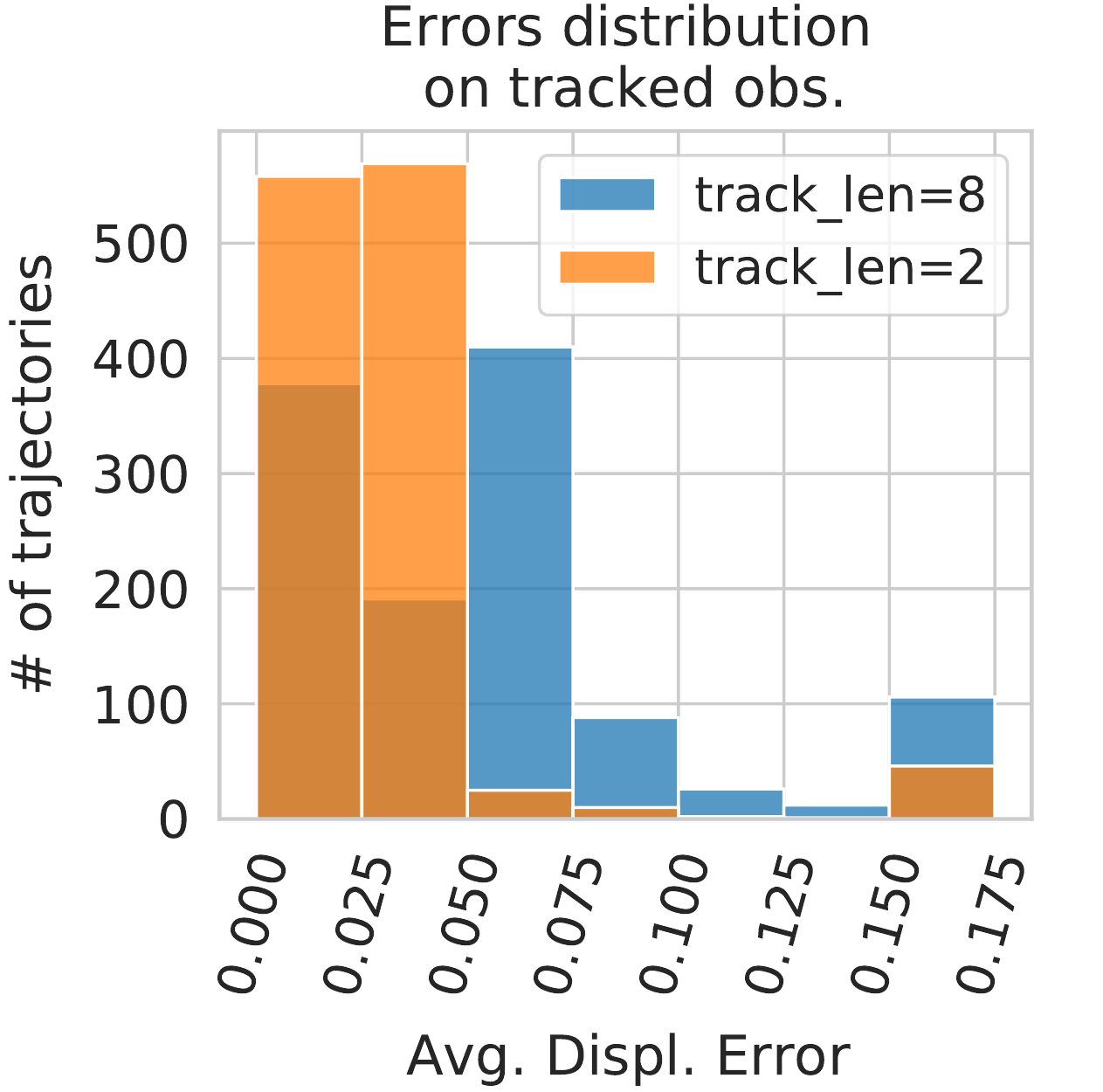}\label{fig:tracked_ades}}\hfil
  \subfloat[]{\includegraphics[scale=0.323]{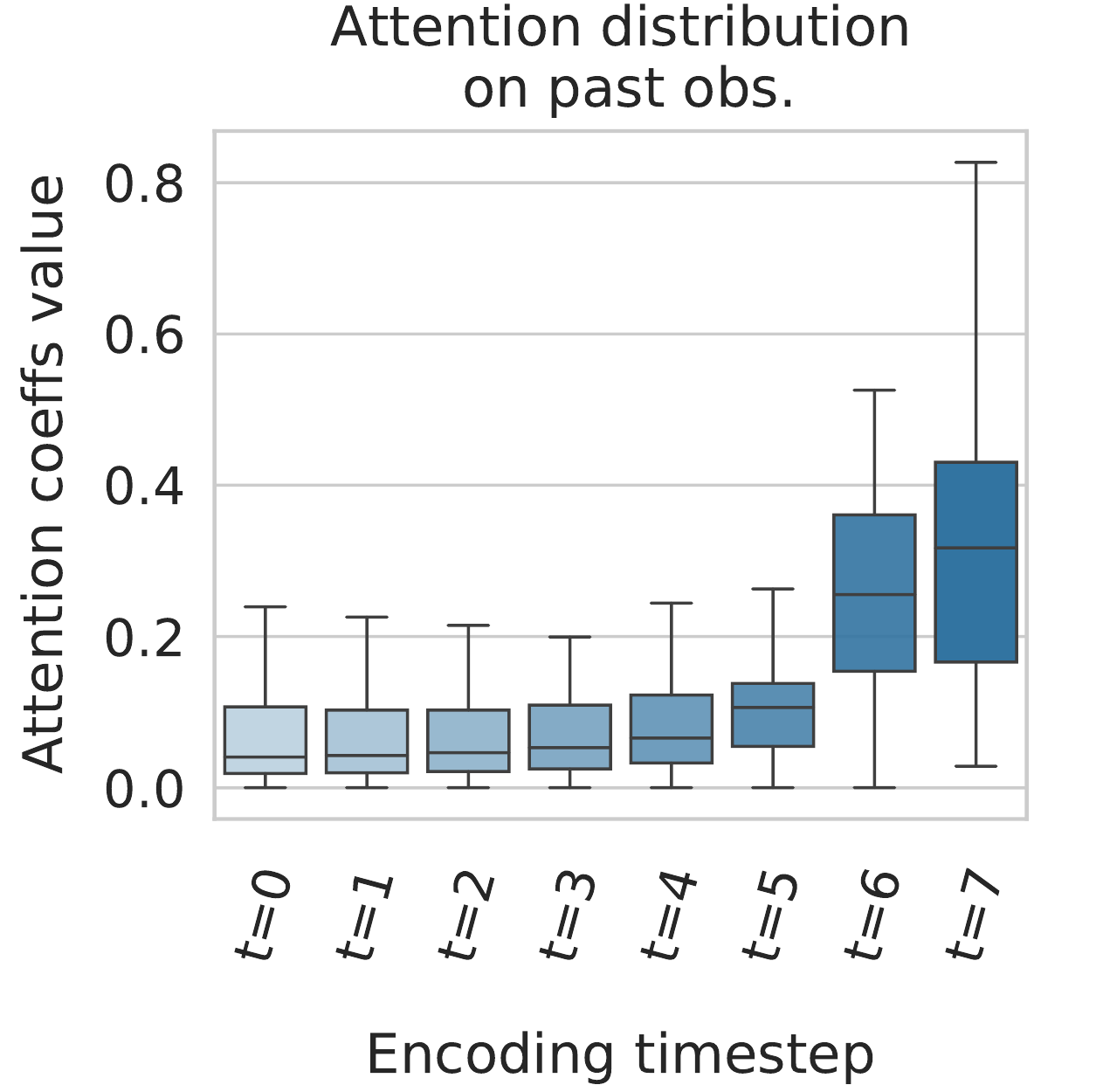}\label{fig:att_coeffs_boxplots}}
  \caption{a) histograms of displacements errors between ground-truth trajectories and their estimations provided by Deep SORT; b) spread of the attention coefficients assigned by the decoder to each encoder state.}
  \label{fig:tracked_ades_and_boxplots}
\end{figure}
\begin{table*}[t]
    \centering
    \addtolength{\tabcolsep}{-0.2pt}
    \begin{tabular}{llcccccccccccccc}
      \toprule
        \textbf{Dataset} & \textbf{Training} & \textbf{obs=2} & \textbf{obs=3} & \textbf{obs=4} & \textbf{obs=5} & \textbf{obs=6} & \textbf{obs=7} & \textbf{obs=8} \\
        \midrule
        \addlinespace
        \multicolumn{1}{c}{\multirow{4}[4]{*}{\makecell{ETH\\UCY}}} & From scratch & 0.56 / 1.12 & 0.51 / 1.08 & 0.48 / 1.01 & 0.47 / 0.90 & 0.46 / 0.96 & 0.45 / 0.95 & \underline{\textit{\textbf{0.43} / \textbf{0.88}}} \\
        \cmidrule{2-9}
        & Variable obs. & 0.64 / 1.33 & 0.63 / 1.31 & 0.61 / 1.28 & 0.62 / 1.28 & 0.62 / 1.29 & 0.63 / 1.31 & 0.64 / 1.31 \\
        & Past generation & 0.50 / 1.06 & 0.47 / 1.01 & 0.46 / 0.98 & 0.46 / 0.96 & 0.45 / 0.95 & 0.45 / 0.91 & - \\
        & \textbf{DTO} & \textbf{0.46} / \textbf{0.93} & \textbf{0.44} / \textbf{0.91} & \textbf{0.43} / \textbf{0.88} & \textbf{0.43} / \textbf{0.88} & \textbf{0.43} / \textbf{0.88} & \textbf{0.43} / \textbf{0.88} & 0.43 / 0.91 \\
        \addlinespace
        \midrule
        \addlinespace
        \multicolumn{1}{c}{\multirow{4}[4]{*}{Lyft}} & From scratch & 0.31 / 0.56 & 0.30 / 0.60 & 0.28 / 0.58 & 0.27 / 0.57 & 0.26 / 0.60 & 0.26 / 0.58 & \underline{\textit{\textbf{0.24} / \textbf{0.53}}} \\
        \cmidrule{2-9}
        & Variable obs. & 0.43 / 0.83 & 0.41 / 0.76 & 0.41 / 0.72 & 0.36 / 0.67 & 0.36 / 0.67 & 0.43 / 0.73 & 0.57 / 0.87 \\
        & Past generation & 0.36 / 0.67 & 0.35 / 0.72 & 0.36 / 0.81 & 0.36 / 0.73 & 0.32 / 0.70 & 0.28 / 0.64 & - \\
        & \textbf{DTO} & \textbf{0.27} / \textbf{0.55} & \textbf{0.26} / \textbf{0.52} & \textbf{0.25} / \textbf{0.52} & \textbf{0.24} / \textbf{0.54} & \textbf{0.25} / \textbf{0.54} & \textbf{0.24} / \textbf{0.55} & 0.25 / 0.55 \\
        \addlinespace
        \bottomrule
    \end{tabular}
    \caption{Comparison (ADE/FDE) between different training strategies; all methods are trained and tested on the same number of time steps, reported in the header. Best results are in bold. The distillation teacher is in underlined italic.}
    \label{tab:addr_length_shift}
\end{table*}
Results reported above suggest that information about future positions can be often recovered by looking solely at the most recent observations. This finding is also investigated by Schöller \etal ~\cite{scholler2020cvm}, who report that forecasting methods retain only partial input data. Furthermore, Becker \etal~\cite{becker2018red} show that the contribution of the latest time step is $80.3$\% while for the second-latest is only $8.3$\%.

To verify if this behaviour also affects our spatio-temporal transformer, Fig.~\ref{fig:att_coeffs_boxplots} reports an analysis of the values assumed by the coefficients in the encoder-decoder self-attention, \ie, the coefficients that represent the contribution of each encoder state to the decoding of future positions. Similarly to~\cite{scholler2020cvm}, we observe that, while earlier steps exert an (albeit small) influence, subsequent states provide a higher contribution. In this regard, we conjecture that the robustness of DTO resides in how the model handles the earlier information: at training time, initial time steps are not drastically discarded (as would happen when training from scratch on fewer steps) but, instead, the student learns to dispense with their limited informative content.

\subsection{On the \quotationmarks{length-shift} problem}
\label{subsec:lengthshift}
\begin{figure}
    \centering
    \includegraphics[width=1.0\columnwidth]{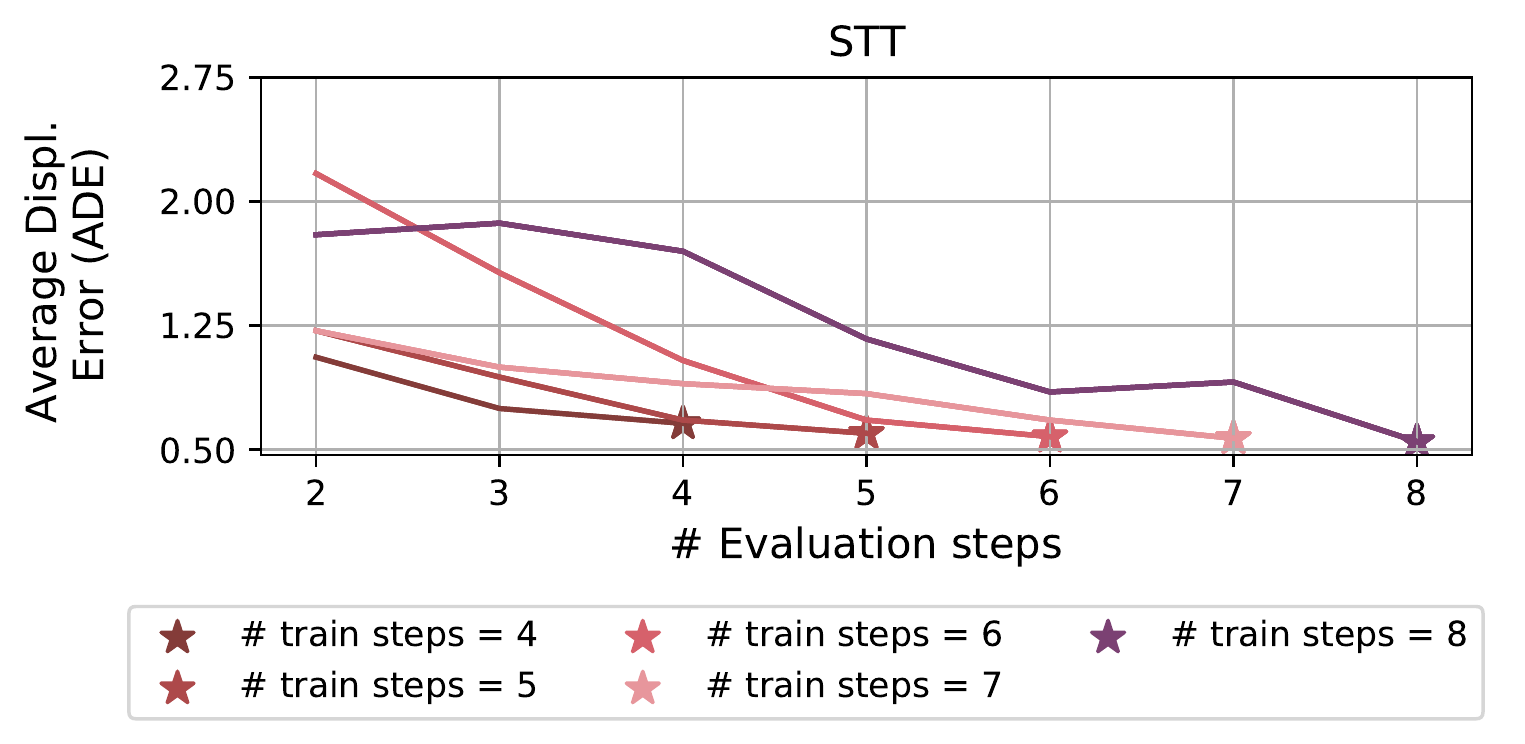}
    \caption{Performance varying the number of observations during evaluation (ETH). The optimum always occurs when training and test conditions match.}
    \label{fig:length_shift}
\end{figure}

We also argue that exploiting longer sequences overly binds the model to the amount of data considered at training time. To prove our intuition, we investigate how models behave when the number of input time steps changes at evaluation time: as shown in Fig.~\ref{fig:length_shift}, reducing the number of past observations results in a sudden and huge performance drop, even for small variations as removing a single time step. 
This behaviour -- which we refer as \quotationmarks{\textbf{length-shift problem}} -- is a common trait among different splits ($8$--$12$, $7$--$12$, etc.) and architectures~\footnote{Suppl. materials report assessments also on V-RNN and SR-LSTM}. This issue could reduce the applicability of these models when limited or partial annotations are available. For this reason, in the following, we explore several strategies that attempt to mitigate this issue: among all, DTO appears the most promising paradigm. 

\medskip
\tinytit{Addressing the length-shift problem} The na\"ive approach for dealing with this problem is to directly train a forecasting model using fewer time steps (\ie, the same number of observations expected at inference time). However, as reported in Sec.~\ref{sec:comparisons} and Tab.~\ref{tab:addr_length_shift}, this choice does not allow the model to extract valuable motion patterns.
\begin{figure*}[t]
  \centering
  \subfloat[\textit{deathCircle\_$0$}]{\includegraphics[scale=0.20]{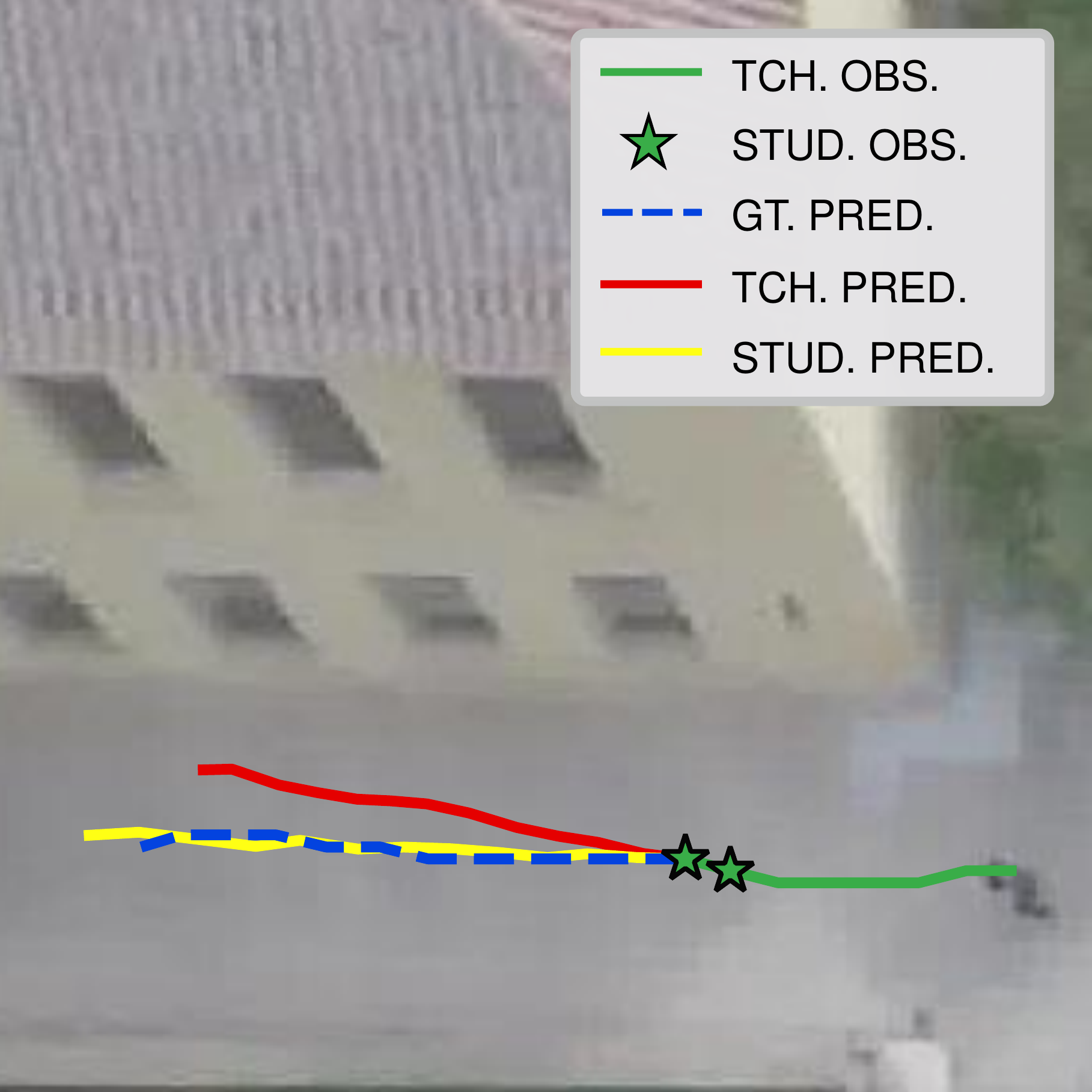}\label{fig:qualitatives_1}}\qquad
  \subfloat[\textit{bookstore\_$0$}]{\includegraphics[scale=0.20]{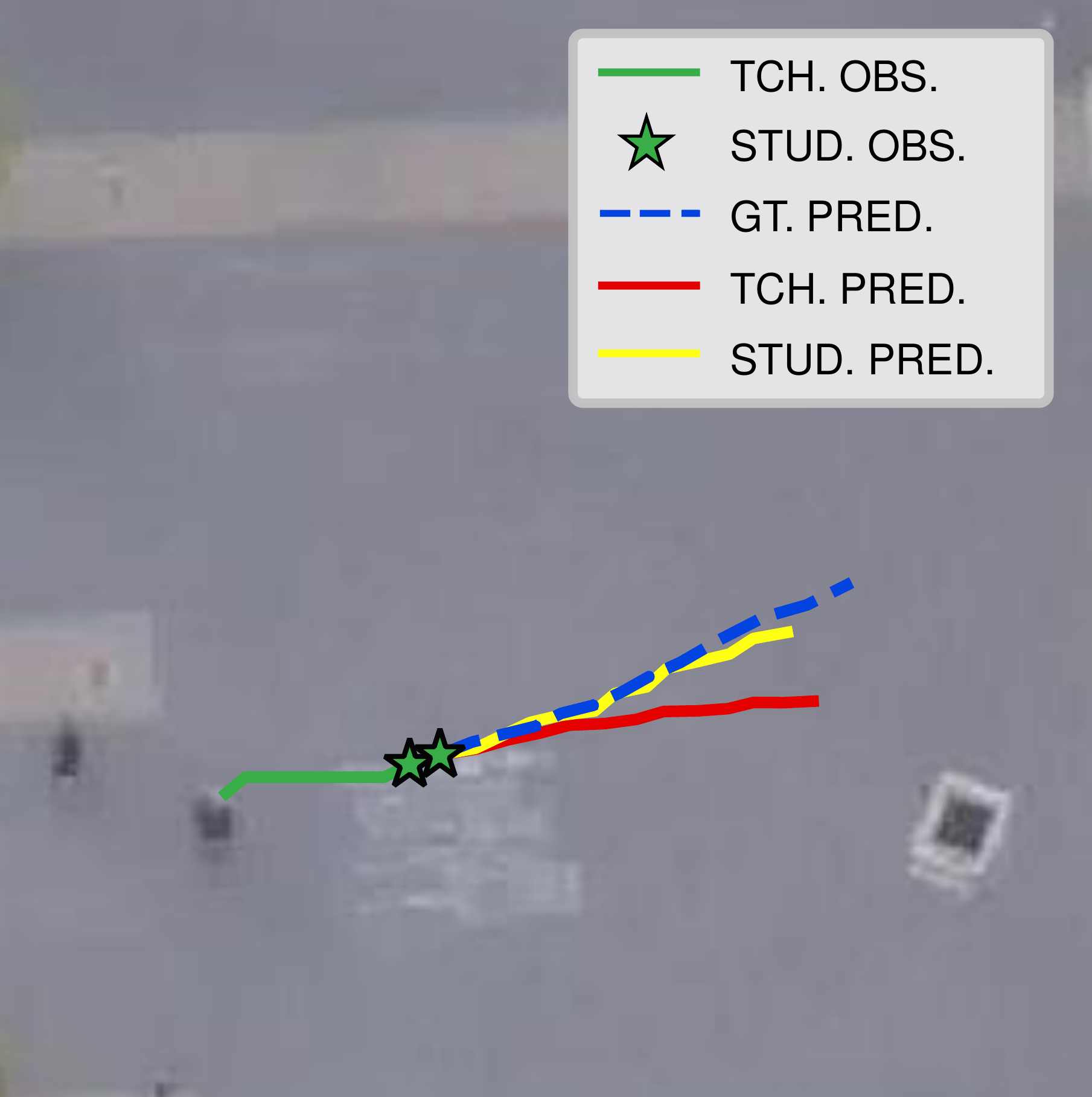}\label{fig:qualitatives_2}}\qquad
   \subfloat[\textit{deathCircle\_$0$}]{\includegraphics[scale=0.20]{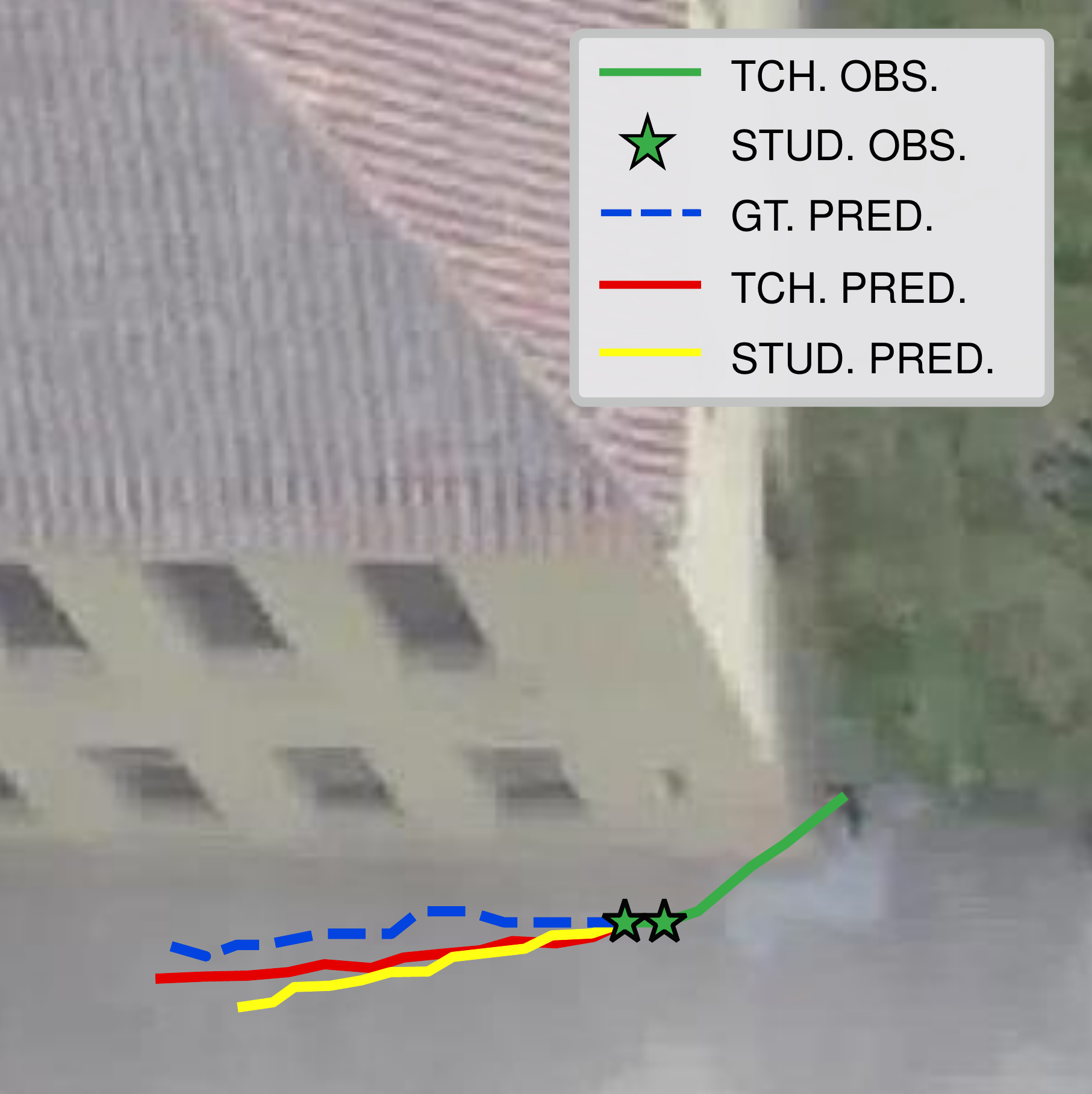}\label{fig:qualitatives_3}}\qquad
  \subfloat[\textit{deathCircle\_$1$}]{\includegraphics[scale=0.20]{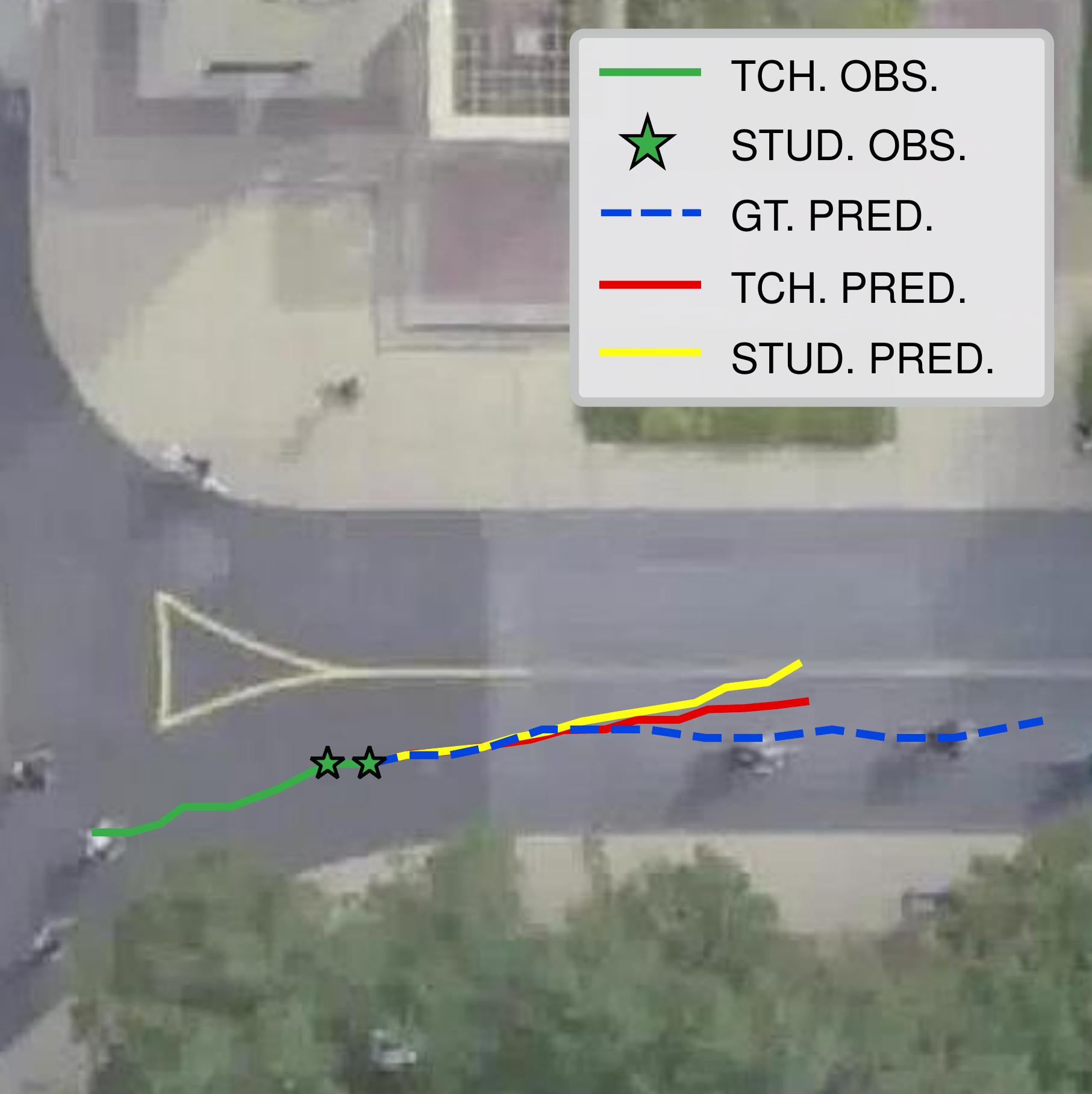}\label{fig:qualitatives_4}}\qquad
  \caption{A qualitative comparison between predicted trajectories generated by our teacher and its $2$--obs distilled student (on Stanford Drone Dataset). In (a) and (b), our student generates more realistic samples, while in (c) and (d), due to more complex dynamics, few observations are not enough to forecast positions close to the ground-truth trajectories.}
  \label{fig:qualitatives}
\end{figure*}
To this end, we explore a second strategy, training a single instance of our STT with a variable number of observations (from $2$ to $8$ time steps). As reported in Tab.~\ref{tab:addr_length_shift} (\textit{Variable observations}), this strategy brings no benefits: we conjecture that the model learns an average set of motion features, thus granting predictions that are less sensitive to changes in the number of time steps; however, it is far from extracting the set of features that is optimal for each specific input length.

A third approach (\textit{Past generation}) reckons on an auxiliary network to fill the input sequence with a set of generated observations: namely, when the number of observations is less than the one used at training time, we employ a secondary model that predicts the missing part of the input trajectory, which is then concatenated to available positions and fed to the primary forecasting model. This represents a step forward, but still delivers unsatisfactory results: we conjecture that the main limitation of this approach regards the amount of noise injected by the auxiliary module, which then spreads to the model that generates future locations.

Finally, we found particularly beneficial the supervision inherent with our distillation strategy. While the student can collect novel motion patterns from few observations as if it was trained from scratch, it also inherits the broader knowledge lying in teacher's activations. This strategy leads to a remarkable gain in performance, outperforming other solutions and reaching teacher's results ($8$ -- $12$).
\subsection{On the \quotationmarks{domain shift} problem -- Knowledge Transfer}
Regarding the generalization capabilities of the student, we discuss here its higher degree of robustness to \textbf{domain-shifts} (\ie, a change in the underlying data distribution between training and test sets). We expect that exploiting only a few observations limits an excessive specialization on the dataset-specific statistics, thus granting a superior strength to distribution shifts. We validate our claim by investigating the following experimental setting: we use SDD as training set and then test both teacher and student networks on novel scenarios, embodied by the test sets of ETH and Lyft.

As reported in Tab.~\ref{tab:knowledge_transf}, the presence of DTO favours knowledge transfer and outperforms its distillation-less counterpart. Moreover, in some cases, this strategy even outperforms the results provided by the upper bound, \ie, when there is a match between source and target datasets (\eg, ETH $\rightarrow$ ETH). On the one hand, we conjecture that this is due to the fact that Stanford Drone collects more representative instances of motion dynamics: indeed, the complexity of the learned motion patterns eases the network effort when evaluated on more straightforward scenarios, such as ETH. On the other hand, our framework proves to be beneficial against shifts, thus providing a solution that addresses scenarios with limited available labels.
\input{tables/transfer}
\subsection{Qualitative analysis}
In some cases (Fig.~\ref{fig:qualitatives_1} and~\ref{fig:qualitatives_2}), the gap occurring between our networks does not impact the corresponding predictions. In presence of complex dynamics, \eg, pronounced turns (see Fig.~\ref{fig:qualitatives_3} and~\ref{fig:qualitatives_4}), the student incurs some issues: here, observing only few positions does not provide enough information to grasp such sophisticated dynamics.

%% file: tables/results.tex
\begin{table*}[t]
\centering
\begin{tabular}{lcccccc}
\toprule
& \textbf{ETH} & \textbf{Hotel} & \textbf{Univ} & \textbf{Zara-1} & \textbf{Zara-2} & \textbf{AVG} \\
\midrule
CVM~\cite{scholler2020cvm} & 1.07 / 2.28 & 0.32 / 0.61 & 0.52 / 1.17 & 0.43 / 0.95 & 0.32 / 0.72 & 0.53 / 1.15\\
ST-GAT 1V-1~\cite{huang2019stgat} & 0.69 / 1.36 & 0.44 / 0.90 & 0.58 / 1.23 & 0.47 / 1.02 & 0.40 / 0.86 & 0.52 / 1.07\\
Ind-TF~\cite{giuliari2020transformer} & 0.60 / 1.25 & 0.27 / 0.50 & 0.64 / 1.23 & 0.57 / 1.09 & 0.42 / 0.81 & 0.50 / 0.96 \\
SR-LSTM~\cite{zhang2019sr} & 0.63 / 1.25 & 0.37 / 0.73 & \textbf{0.51} / \textbf{1.10} & 0.41 / 0.90 & 0.32 / \textbf{0.70} & 0.45 / 0.94\\
STAR~\cite{yu2020spatio} & 0.56 / 1.11 & 0.26 / 0.50 & 0.52 / 1.13 & \textbf{0.40} / \textbf{0.89} & \textbf{0.31} / 0.71 & \textbf{0.41} / \textbf{0.87}\\
\midrule
\textbf{STT} (8 obs)& \textbf{0.54} / \textbf{1.10} & \textbf{0.24} / \textbf{0.46} & 0.57 / 1.15 & 0.45 / 0.94 & 0.36 / 0.77 & 0.43 / 0.88\\
\textbf{STT} (2 obs)& 0.72 / 1.45 & 0.48 / 0.48 & 0.53 / 1.09 & 0.64 / 1.21 & 0.44 / 0.88 & 0.57 / 1.12 \\
\textbf{STT + DTO} (2 obs) & 0.62 / 1.22 & 0.29 / 0.56 & 0.58 / 1.14 & 0.45 / 0.98 & 0.34 / 0.74 & 0.46 / 0.93 \\
\bottomrule
\end{tabular}
\caption{Comparisons (in terms of ADE/FDE) on ETH/UCY. Our teacher network (STT) trained according to the standard protocol shows comparable results w.r.t the competitors, while our student network (STT + DTO) shows similar performance despite its knowledge gap.}
\label{tab:eth_adefde_sota_comparison}
\end{table*}

%% file: tables/vrnn.tex
\begin{table}[t]
    \centering
    \begin{tabular}{lcc}
        \toprule
            & \textbf{Sampling strategy} & \textbf{ADE} / \textbf{FDE} \\
        \midrule
            VRNN-1 & one sample & 0.73 / 1.49 \\
            VRNN-20 & argmin $\text{KL}(q \parallel p)$ & 0.75 / 1.51 \\
            VRNN-20 & argmin $\text{MSE}(\cdot, \text{GT})$ & \textbf{0.58} / \textbf{1.17} \\
        \midrule
            STT (ours) & one sample & \underline{0.63} / \underline{1.26} \\
        \bottomrule
    \end{tabular}
    \caption{Comparison between our approach and the V-RNN~\cite{chung2015recurrent}.}
    \label{tab:stochastic_vs_deterministic_tracked}
\end{table}

%% file: tables/tracking.tex
\begin{table*}
\centering
\begin{tabular}{lccccccccc}
    \toprule
\multicolumn{1}{c}{\multirow{2}[2]{*}{\textbf{Scene}}} & \multicolumn{2}{c}{\textbf{Ground Truth}} && \multicolumn{4}{c}{\textbf{Tracked trajectories}} \\
\cmidrule{2-3} \cmidrule{5-8} & STT (8 obs) & DTO (2 obs) && STT (8 obs) & STT (2 obs) & CVM (2 obs) & DTO (2 obs) \\
\midrule
bookstore & \textbf{0.48 / 0.97} & 0.49 / 0.95 && 0.58 / 1.08 & 0.54 / 1.01 & 0.55 / 1.02 & \textbf{0.53 / 0.99}\\
nexus & \textbf{0.64 / 1.26} & 0.72 / 1.39 && 1.38 / 2.10 & 1.33 / 2.07 & 1.38 / 2.15 & \textbf{1.29 / 2.03}\\
deathCircle & \textbf{0.76 / 1.55} & 0.85 / 1.74 && 0.99 / 1.83 & 0.97 / 1.83 & 0.99 / 1.86 & \textbf{0.94 / 1.82}\\
gates & \textbf{0.75 / 1.63} & 0.82 / 1.72 && 1.20 / 2.15 & 1.00 / 1.94 & 1.10 / 1.97 & \textbf{0.94 / 1.84}\\
hyang & \textbf{0.37 / 0.80} & 0.38 / 0.78 && \textbf{0.39 / 0.82} & 0.48 / 0.95 & 0.41 / 0.85 & 0.46 / 0.89\\
coupa & 0.20 / 0.40 & \textbf{0.20 / 0.38} && 0.28 / 0.49 & 0.21 / 0.41 & 0.26 / 0.44 & \textbf{0.20 / 0.38}\\
\midrule
overall & \textbf{0.55 / 1.12} & 0.60 / 1.19 && 0.84 / 1.46 & 0.80 / 1.40 & 0.84 / 1.40 & \textbf{0.77 / 1.37}\\
    \bottomrule
 \end{tabular}
\caption{On the SDD's scenarios, a comparison (ADE / FDE) between teacher (STT) and student (STT + DTO) on both ground-truth and tracked trajectories.}
\label{tab:gt_tracked_stt_vs_dto}
\end{table*}

%% file: tables/transfer.tex
\begin{table}[t]
    \centering
    \addtolength{\tabcolsep}{-2.3pt}
    \begin{subtable}{\linewidth}
    \begin{tabular}{ccccccccc}
            \toprule
             \multicolumn{1}{c}{\multirow{2}[2]{*}{
             \shortstack{\textbf{Train.}\\ \textbf{Set}}}}&\multicolumn{1}{c}{\multirow{2}[2]{*}{\textbf{DTO}}} & \multicolumn{7}{c}{\textbf{\# of training/evaluation obs}}\\
                \cmidrule{3-9}
                & & 2 & 3 & 4 & 5 & 6 & 7 & 8 \\
            \midrule
                ETH & \xmark & 0.72 & 0.68 & 0.66 & 0.60 & 0.58 & 0.57 & 0.54 \\
                SDD & \xmark & 0.71 & 0.59 & \textbf{0.57} & 0.58 & 0.57 & 0.57 & 0.55 \\
                SDD & \cmark & \textbf{0.66} & \textbf{0.57} & 0.58 & \textbf{0.54} & \textbf{0.56} & \textbf{0.54} & \textbf{0.55} \\
            \bottomrule
         \end{tabular}
         \caption{Performance on the test set of ETH.}
    \end{subtable}\\
    \begin{subtable}{\linewidth}
    \begin{tabular}{ccccccccc}
            \toprule
     \multicolumn{1}{c}{\multirow{2}[2]{*}{
             \shortstack{\textbf{Train.}\\ \textbf{Set}}}}&\multicolumn{1}{c}{\multirow{2}[2]{*}{\textbf{DTO}}} & \multicolumn{7}{c}{\textbf{\# of training/evaluation obs}}\\
                \cmidrule{3-9}
                & & 2 & 3 & 4 & 5 & 6 & 7 & 8 \\
            \midrule
                Lyft & \xmark & 0.31 & 0.30 & 0.28 & 0.27 & 0.26 & 0.26 & 0.24 \\
                SDD & \xmark & 0.70 & 0.53 & 0.41 & \textbf{0.41} & 0.44 & 0.46 & 0.41 \\
                SDD & \cmark & \textbf{0.35} & \textbf{0.30} & \textbf{0.30} & 0.42 & \textbf{0.36} & \textbf{0.28} & \textbf{0.34}\\
            \bottomrule
         \end{tabular}
         \caption{Performance on the test set of Lyft.}
    \end{subtable}
     \caption{Results of transfer knowledge among different datasets (ADE). We highlight in bold the highest results obtained in case of dataset-shift (\textit{i.e.}, second and third rows of each of the two tables).}
     \label{tab:knowledge_transf}
\end{table}

%% file: sections/5_conclusion.tex
\section{Conclusion}
This paper proposes an in-depth analysis of the evaluation protocol usually employed to assess trajectory prediction models. We conceive a novel training strategy to train a transformer-based architecture to deal with scenarios where only a few observations are available. Our teacher-student paradigm reduces the information gap experienced by the student, thus providing a practical and viable inference scheme for on-line scenarios. We also investigate issues that could emerge at inference time. Our experiments suggest that our strategy also enables a better knowledge transfer capability across different training scenarios.

\medskip
\tinytit{Acknowledgments}~Funded by the PREVUE \quotationmarks{PRediction of activities and Events by Vision in an Urban Environment} project (CUP E94I19000650001), PRIN National Research Program, Italian Ministry for Education, University and Research (MIUR).